\documentclass[10pt,twocolumn,letterpaper]{article}

\usepackage[pagenumbers]{cvpr} 

\usepackage{times}
\usepackage{graphicx}
\usepackage{amsmath}
\usepackage{amssymb}

\usepackage{cite}
\usepackage{multirow}
\usepackage{verbatim}
\usepackage{mathtools}
\usepackage{xcolor}
\usepackage{booktabs}

\usepackage[accsupp]{axessibility}

\definecolor{cvprblue}{rgb}{0.21,0.49,0.74}
\usepackage[pagebackref,breaklinks,colorlinks,citecolor=cvprblue]{hyperref}

\newcommand{\bA}{\ensuremath{{\mathbf{A}}}}
\newcommand{\bD}{\ensuremath{{\mathbf{D}}}}
\newcommand{\bG}{\ensuremath{{\mathbf{G}}}}
\newcommand{\bR}{\ensuremath{{\mathbf{R}}}}

\newcommand{\bI}{\ensuremath{{\mathbf{I}}}}

\newcommand{\bQ}{\ensuremath{{\mathbf{Q}}}}
\newcommand{\bK}{\ensuremath{{\mathbf{K}}}}

\newcommand{\bM}{\ensuremath{{\mathbf{M}}}}

\newcommand{\bS}{\ensuremath{\mathbf{S}}}
\newcommand{\bW}{\ensuremath{\mathbf{W}}}

\newcommand{\ba}{\ensuremath{{\mathbf{a}}}}
\newcommand{\bb}{\ensuremath{{\mathbf{b}}}}

\newcommand{\bm}{\ensuremath{{\mathbf{m}}}}
\newcommand{\bq}{\ensuremath{{\mathbf{q}}}}
\newcommand{\br}{\ensuremath{{\mathbf{r}}}}

\newcommand{\cL}{\ensuremath{{\mathcal{L}}}}

\usepackage[capitalize]{cleveref}
\crefname{section}{Sec.}{Secs.}
\Crefname{section}{Section}{Sections}
\Crefname{table}{Table}{Tables}
\crefname{table}{Tab.}{Tabs.}

\def\paperID{1076} 
\def\confName{CVPR}
\def\confYear{2024}

\title{Masked Spatial Propagation Network for Sparsity-Adaptive Depth Refinement}

\author{Jinyoung Jun\\
Korea University\\
{\tt\small jyjun@mcl.korea.ac.kr}
\and
Jae-Han Lee\\
Gauss Labs Inc.\\
{\tt\small jaehanlee@mcl.korea.ac.kr}
\and
Chang-Su Kim\thanks{Corresponding author.}\\
Korea University\\
{\tt\small changsukim@korea.ac.kr}
}

\begin{document}
\maketitle

\begin{abstract}
The main function of depth completion is to compensate for an insufficient and unpredictable number of sparse depth measurements of hardware sensors. However, existing research on depth completion assumes that the sparsity --- the number of points or LiDAR lines --- is fixed for training and testing. Hence, the completion performance drops severely when the number of sparse depths changes significantly. To address this issue, we propose the sparsity-adaptive depth refinement (SDR) framework, which refines monocular depth estimates using sparse depth points. For SDR, we propose the masked spatial propagation network (MSPN) to perform SDR with a varying number of sparse depths effectively by gradually propagating sparse depth information throughout the entire depth map. Experimental results demonstrate that MPSN achieves state-of-the-art performance on both SDR and conventional depth completion scenarios. Codes are
available at \url{https://github.com/jyjunmcl/MSPN_SDR}
\end{abstract}

\section{Introduction}
\label{sec:introduction}

Image-guided depth completion is a task to estimate a dense depth map using an RGB image with sparse depth measurements; it fills in unmeasured regions with estimated depths. It is useful because many depth sensors, \eg LiDAR and ToF cameras, provide sparse depth maps only. With the recent usage of depth information in autonomous driving \cite{geiger2012we} and various 3D applications \cite{ye2011accurate, Izadinia2017CVPR, Liu2018CVPR, shih20203d}, depth completion has become an important research topic.

Recently, with the success of deep neural networks, learning-based methods have shown significant performance improvement by exploiting massive amounts of training data \cite{ma2018sparse}. They attempt to fuse multi-modal features like surface normal \cite{xu2019depth, qiu2019deeplidar} or provide repetitive image guidance \cite{yan2022rignet}. Especially, affinity-based spatial propagation methods have been widely studied \cite{cheng2018depth, cheng2020cspn++, park2020non, lin2022dynamic, zhang2023completionformer}.

\begin{figure}[!t]
  \centering
   \includegraphics[width=\linewidth]{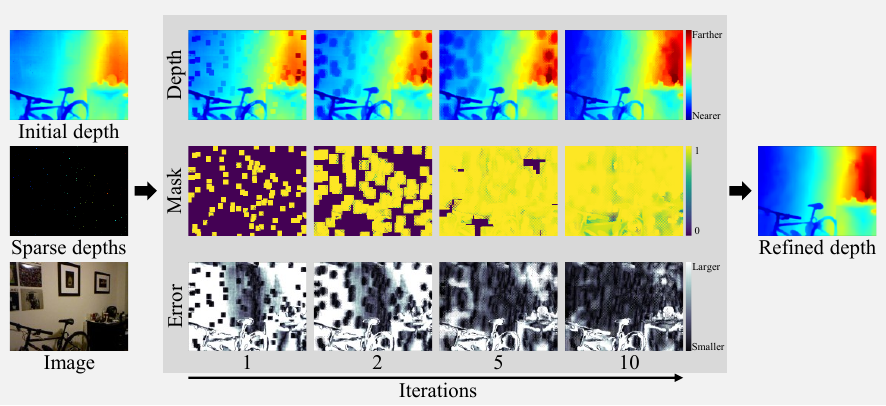}
   \caption{Illustration of the masked spatial propagation process in the proposed MSPN. The initial mask is obtained from sparse depths, assigned 1 if depth values are present and 0 otherwise. For easier comparison, error maps are also provided for each depth map, in which brighter pixels correspond to larger errors. MSPN updates depth maps and masks gradually to generate the final refined depth map.}
   \label{fig:intro}
   \vspace{-0.2cm}
\end{figure}

The main function of depth completion is to compensate for the limitations of existing depth sensors, but conventional research on depth completion assumes that the sparsity is fixed for training and testing. However, in practice, sparsity changes significantly, for it is difficult to measure the depths of transparent regions, mirrors, and black objects. Sensor defects also affect the number of measurements. Besides, conventional spatial propagation methods \cite{cheng2018depth, cheng2020cspn++, park2020non, lin2022dynamic, zhang2023completionformer} refine the depths of all pixels simultaneously, regardless of the locations of sparse depth measurements. Therefore, erroneous depths can propagate during the refinement when only a few sparse depths are available.

In this paper, we develop a sparsity-adaptive depth refinement (SDR) framework, which refines monocular dense depth estimates adaptively according to the sparsity of depth measurements. Also, we propose the masked spatial propagation network (MSPN) to propagate the information from sparse depth points to unmeasured regions. First, an off-the-shelf monocular depth estimator is used to estimate an initial depth map from an input RGB image. Next, a guidance network generates guidance features using the input image, the sparse depths, and the initial depth map. Finally, using the guidance features, the proposed MSPN performs iterative refinement to obtain a refined depth map, as shown in Figure \ref{fig:intro}. The proposed SDR framework can be trained with a variable number of sparse depths, making it more suitable for real-world applications. In addition, the proposed MSPN performs significantly better than conventional methods in the SDR scenario by generating an adaptive propagation mask according to sparse measurements. Moreover, MSPN provides state-of-the-art performance on the conventional depth completion on NYUv2 \cite{silberman2012indoor} and KITTI \cite{geiger2012we} datasets.

This paper has the following contributions:
\begin{itemize}
\itemsep0em
  \item We develop the SDR framework, which refines monocular depth estimates using a variable number of sparse depth measurements.
  \item For SDR, we propose MSPN to handle a various number of sparse depths by gradually propagating sparse depth information.
  \item MSPN provides state-of-the-art performance on both SDR and conventional depth completion scenarios.
\end{itemize}

\section{Related Work}
\label{sec:related_work}
\subsection{Monocular depth estimation}
The goal of monocular depth estimation is to infer the absolute distance of each pixel in a single image from the camera. Traditional approaches are based on assumptions about the 3D scene structure --- superpixels \cite{saxena2008make3d}, block world \cite{gupta2010eccv}, or line segments and vanishing points \cite{gupta2010nips}. However, such assumptions are invalid for small objects or due to color ambiguity, making monocular depth estimation an ill-posed problem. With the advance of CNNs, training-based methods have successfully overcome the ill-posedness of monocular depth estimation. Earlier CNN methods focused on exploring better architectures \cite{eigen2014depth, laina2016deeper, xu2017multi, heo2018monocular, chen2019structure} or loss functions \cite{eigen2015predicting, chen2016single, laina2016deeper, hu2019revisiting, lee2020multi} for more effective training. Recently, as transformer and self-attention \cite{vaswani2017attention} have been applied to diverse vision tasks \cite{dosovitskiy2020image}, transformers for monocular depth estimation have also been developed \cite{yuan2022newcrfs}.

On the other hand, training strategies to utilize 3D information have been attempted. For example, merging depth maps in frequency domain \cite{lee2018single}, enforcing high-order geometric constraints by defining virtual normal \cite{yin2019enforcing}, and computing depth attention volume which considers pixel-to-image attention \cite{huynh2020guiding}, have been attempted. Also, ordinal regression \cite{fu2018deep}, or planar coefficient estimation \cite{patil2022p3depth} have been attempted.

\subsection{Image-guided depth completion}
Whereas monocular depth estimation uses only a single image as input, image-guided depth completion focuses on filling depths of unmeasured regions in an image when sparse depth measurements are provided. Significant progress has been made after the emergence of learning-based methods, overcoming the modality gap between image and depth. Early methods on depth completion concatenate images and sparse depth maps and process them using an encoder-decoder network \cite{ma2018sparse, ma2019self}. Also, attempts have been made to extract image and depth features separately and then leverage them to compensate for the modality gap effectively. For example, multi-branch architecture \cite{hu2021penet, yan2022rignet, rho2022guideformer, nazir2022semattnet} and coarse-to-fine framework \cite{xu2020deformable, liu2021fcfr} have been studied. More specifically, in the coarse-to-fine framework, variants of affinity-based spatial propagation \cite{liu2017learning} are dominant \cite{cheng2018depth, cheng2019learning, cheng2020cspn++, xu2020deformable, park2020non, lin2022dynamic}, since it can be easily applied using various encoder-decoder networks \cite{hu2021penet, nazir2022semattnet, zhang2023completionformer}.

\subsection{Spatial propagation network}
Motivated by anisotropic diffusion \cite{weickert1998anisotropic}, spatial propagation networks (SPNs) refine initial depth estimates using sparse depth measurements \cite{cheng2018depth, cheng2019learning, cheng2020cspn++, xu2020deformable, park2020non, lin2022dynamic}. They can reduce blurring caused by the encoder-decoder structure with varying spatial resolutions. To reconstruct the depth of a pixel based on spatial propagation, reference pixels, and their referring methods should be determined. The original SPN \cite{liu2017learning} uses three adjacent pixels in a neighboring row or column and performs refinement in four directions: up, down, left, and right. Convolution is also used to use neighboring pixels \cite{cheng2019learning, cheng2020cspn++}. Moreover, to refer to distant pixels, deformable convolution \cite{dai2017deformable} is employed in \cite{xu2020deformable, park2020non, lin2022dynamic}.

While these SPN methods show impressive performance, they do not take the sparsity of available depths into account, which is essential in practical applications. For example, a network trained with 500 sparse depth points is less effective when the number of sparse depth points changes significantly since the initial depth map is estimated based on input sparse depths. In contrast, in this paper, we define a propagation mask, which is updated together with a depth map during the refinement. An initial depth map is first updated using only sparse depth points, and then a wider area is refined based on the propagation mask. As a result, the proposed MSPN can refine monocular depths effectively using a variable number of sparse depths.

\subsection{Transformer and self-attention}
After the success of transformers \cite{vaswani2017attention, devlin2018bert} in natural language processing, extensive attempts have been made to apply transformers to vision tasks as well. Self-attention is a key component in transformers, which highlights and extracts important features from input. Recently, many transformers have demonstrated even better performances than CNNs in vision tasks \cite{dosovitskiy2020image, liu2021swin, ranftl2021vision, fang2022msg, chen2021regionvit, wang2021pyramid, hassani2023neighborhood}. Moreover, there have been researches for cross-attention between different types of features \cite{carion2020end, cheng2021per, jaegle2021perceiver}, which can combine different modalities. In this paper, we develop an attention-based refinement algorithm for depth completion. Whereas the conventional spatial propagation performs refinement via convolutions, the proposed MSPN refines the depth map using reliable pixels based on the masked attention strategy.

\begin{figure*}[!t]
  \centering
   \includegraphics[width=\linewidth]{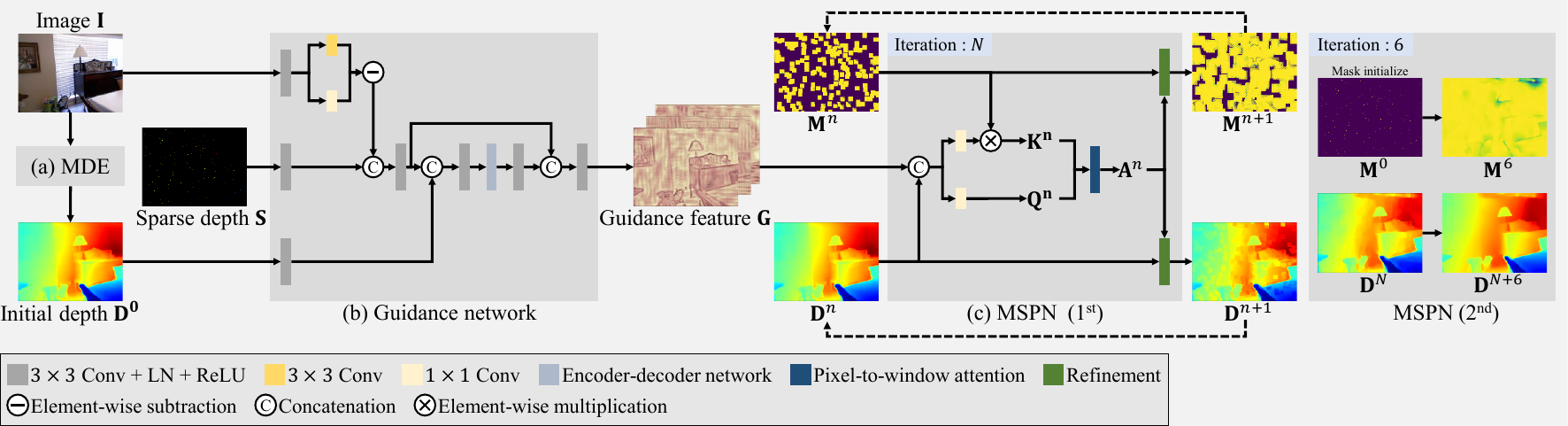}
   \caption{An overview of the proposed SDR framework.}
   \label{fig:overview}
\end{figure*}

\section{Proposed Algorithm}
\label{sec:proposed_algorithm}

\subsection{Conventional depth completion}
Let $\bI \in \mathbb{R}^{3 \times H \times W}$ be an image and $\bS \in \mathbb{R}^{H \times W}$ record its sparse depth measurements. The goal of depth completion is to estimate a dense depth map $\bD \in \mathbb{R}^{H \times W}$ using $\bI$ and $\bS$. However, it is challenging due to the different modalities between color and depth. Moreover, while $\bS$ provides the accurate depths for measured pixels, it does not provide any information on unmeasured regions. Therefore, previous methods \cite{cheng2018depth, cheng2020cspn++, park2020non, lin2022dynamic, zhang2023completionformer} generate a guidance feature $\bG \in \mathbb{R}^{C \times H \times W}$ that compensates for the modality gap, as well as an initial depth prediction $\bD^0$;
\begin{equation}
(\bG, \bD^0) = \theta(\bI, \bS).
\label{eq:guidance_conventional}
\end{equation}
where $\theta$ denotes a multi-head network that estimates both $\bD^0$ and $\bG$. Then, $\bD^0$ is refined $N$ times to obtain the final depth map $\bD^N$
\begin{align}
    \bD^{n+1} = \phi(\bD^n, \bG), \quad n = 0, \dots, N-1,
\end{align}
where $\phi$ is a spatial propagation network.

\subsection{Sparsity-adaptive depth refinement}
The conventional depth completion methods \cite{cheng2018depth, cheng2020cspn++, park2020non, lin2022dynamic, zhang2023completionformer} are trained with a fixed number of sparse depths. In practice, however, external factors often change the number of sparse depths, making the conventional methods less effective. In other words, the number of available depths in $\bS$ in \eqref{eq:guidance_conventional} is variant.

To address this problem, we propose the SDR framework, which consists of three networks: (a) a monocular depth estimator (MDE) that estimates an initial depth map, (b) a guidance network that combines different modality features, and (c) MSPN that recursively updates depth maps and masks. Figure \ref{fig:overview} illustrates an overview of the proposed SDR. Note that we may use any off-the-shelf MDE. In this paper, for MDE, we use the pre-trained parameters of the Jun \etal's algorithm \cite{jun2023versatile} to generate $\bD^0$ and exclude $\bS$ from the generation process. Instead, we use $\bD^0$ to generate $\bG$ as follows:
\begin{equation}
\bG = \Theta(\bI, \bS, \bD^0)
\label{eq:guidance_proposed}
\end{equation}
where $\Theta$ denotes the proposed guidance network.

Next, we refine $\bD^0$ using $\bG$, based on the sparse depth locations in $\bS$. The spatial propagation methods in \cite{cheng2020cspn++, park2020non} refine all pixels simultaneously. Thus, erroneous depth values can propagate to neighbors, especially when only a few sparse depths are available. Instead of refining all pixels in a depth map at once, we define a propagation mask and gradually update the depth map from the sparse depth points. We perform the recursive refinement on both $\bD^n$ and $\bM^n$,
\begin{align}
    (\bD^{n+1}, \bM^{n+1}) = \Phi(\bD^n, \bM^n, \bG), \, n = 0, \dots, N-1,
\end{align}
where $\Phi$ denotes the proposed MSPN. The initial propagation mask $\bM^0 \in \mathbb{R}^{H \times W}$ is defined as
\begin{equation}
\bM^0 = \Xi(\bS)
\label{eq:initial_mask}
\end{equation}
where $\Xi$ denotes an indicator function that outputs 1 for each sparse depth point and 0 otherwise.

\subsection{Guidance network}
As shown in Figure \ref{fig:overview}(b), the guidance network takes $\bI$, $\bS$ and $\bD^0$ and generates the guidance feature $\bG$. To implement the guidance network, we adopt the encoder-decoder architecture in U-Net \cite{ronneberger2015u}. The encoder extracts lower-resolution features from input, and the decoder processes the features to yield $\bG$.

\begin{figure*}[!t]
  \centering
   \includegraphics[width=\linewidth]{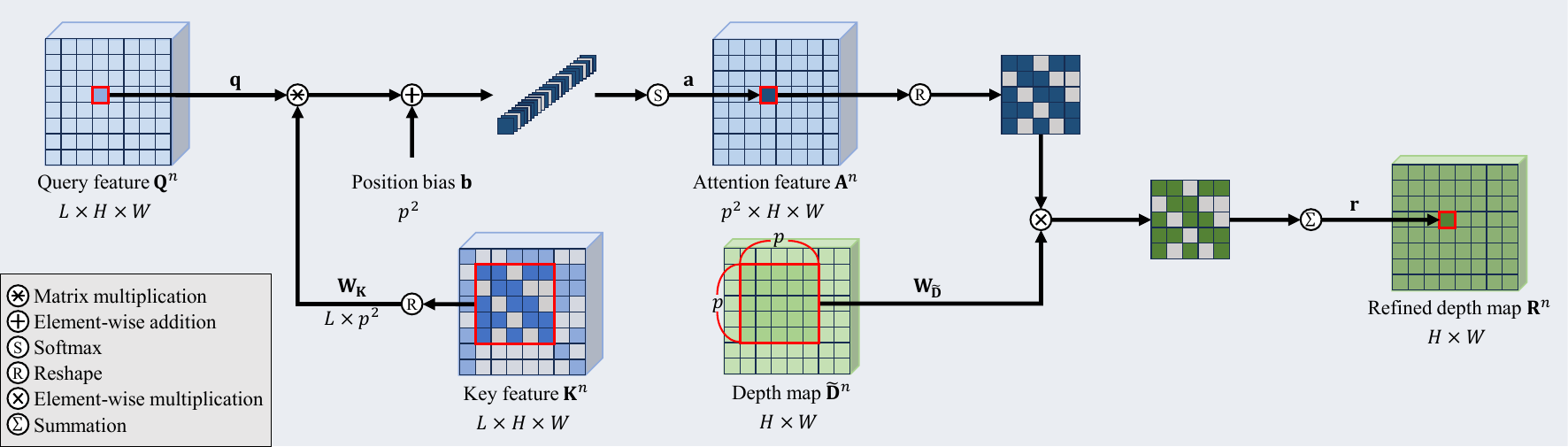}
   \caption{Illustration of the pixel-to-window attention process and generation process of $\bR^n$. $\bM^{n+1}$ is generated in the same manner.}
   \label{fig:mspn}
\end{figure*}

In a depth map, pixels on a large object or background have similar depths and vary gradually. Such depths are low-frequency components. In contrast, depths on small objects or edge regions are high-frequency components. Since sparse depths do not cover an entire image, extracting high-frequency information from the image is crucial for efficient refinement. Hence, we extract high-frequency features by subtracting the results of $1 \times 1$ convolution from those of $3 \times 3$ convolution, as in \cite{xu2020learning}. We extract the high-frequency features from $\bI$, merge them with the features from $\bS$ and $\bD^0$, and feed the result to the encoder. The decoder has five blocks, each consisting of $3 \times 3$ transpose convolution, layer normalization \cite{ba2016layer}, ReLU, and a NAF block \cite{chen2022simple}.

\subsection{Masked spatial propagation network}
Using the output $\bG$ of the guidance network, MSPN updates a depth map $\bD^n$ and a mask $\bM^n$ to $\bD^{n+1}$ and $\bM^{n+1}$, as shown in Figure \ref{fig:overview}(c). Each pixel in $\bD^n$ is refined using its neighboring pixels, while the sparse depth values in $\bS$ remain unchanged. Specifically, we replace the depth values in $\bD^n$ using $\bS$ and generate $\Tilde{\bD}^n$ by
\begin{equation}
    \Tilde{\bD}^n = (1 - \bM^0) \otimes \bD^n + \bM^0 \otimes \bS
\end{equation}
where $\otimes$ denotes the element-wise multiplication.

Next, we determine reference pixels for the refinement and the strength of the refinement. The conventional spatial propagation methods \cite{cheng2018depth, cheng2020cspn++, park2020non, lin2022dynamic} focus on the selection of reference pixels. However, there are much fewer reliable pixels than unreliable ones, so these methods are less effective when only a small number of sparse depths are provided. Instead, we design the masked-attention-based dynamic filter, which computes attention scores between each pixel and its surrounding pixels. We first generate the query feature $\bQ \in \mathbb{R}^{L \times H \times W}$ and the key feature $\bK \in \mathbb{R}^{L \times H \times W}$,
\begin{equation}
    \bQ^n = f_\bQ([\Tilde{\bD}^n, \bG]), \quad \bK^n = f_\bK([\Tilde{\bD}^n, \bG]) \otimes \bM^n
\label{eq:qk}
\end{equation}
where $f_\bQ$ and $f_\bK$ are $1 \times 1$ convolutions followed by layer normalization  \cite{ba2016layer}. Also, $[\cdot]$ denotes the channel-wise concatenation. Since $\Tilde{\bD}^n$ is unrefined, $\bK^n$ is a mixture of reliable and unreliable pixel features. Therefore, we mask unreliable pixel features when computing $\bK^n$ in (\ref{eq:qk}).

Next, we compute the attention scores between $\bQ^n$ and $\bK^n$. Let $\bq \in \mathbb{R}^L$ be a query pixel feature in $\bQ^n$ at location $(i,j)$. Also, let $\bW_\bK \in \mathbb{R}^{L \times p^2}$ denote the $p \times p$ window of the key features in $\bK^n$ centered at $(i,j)$. Note that we compute pixel-to-window attention in order to refine pixel $(i,j)$ using its neighboring pixels. More specifically, the pixel-to-window attention $\ba \in \mathbb{R}^{p^2}$ is computed as
\begin{equation}
    \ba = \text{softmax}(\bq^t \bW_\bK + \bb)
\end{equation}
where $\bb \in \mathbb{R}^{p^2}$ denotes the relative position bias \cite{hassani2023neighborhood} in the $w \times w$ window. By performing the attention on all pixels in $\bQ^n$, the attention feature $\bA^n \in \mathbb{R}^{p^2 \times H \times W}$ is obtained.

Then, using $\bA^n$ and $\Tilde{\bD}^n$, we generate a refined depth map $\bR^n \in \mathbb{R}^{H \times W}$. Let $\bW_{\Tilde{\bD}}$ and $\bW_\bM$ be the $p \times p$ windows in $\Tilde{\bD}^n$ and $\bM^n$, centered at $(i,j)$, respectively. A refined depth pixel $\br$ in $\bR^n$ is obtained by
\begin{equation}
    \br = \sum_{t=1}^{p^2}\ba_t \cdot \bW_{\Tilde{\bD},t}
    \label{eq:rn}
\end{equation}
where the subscript $t$ denotes the $t$th element in the window. Figure \ref{fig:mspn} illustrates the pixel-to-window attention process and the generation process of $\bR^n$.

Finally, depth map $\bD^{n+1}$ at the next iteration step is given by
\begin{equation}
    \bD^{n+1} = (1 - \bM^n) \otimes \Tilde{\bD}^n + \bM^n \otimes \bR^n.
\end{equation}
Also, similarly to (\ref{eq:rn}), mask pixel $\bm^{n+1}$ at the next iteration step is computed by
\begin{equation}
    \bm^{n+1} = \sum_{t=1}^{p^2}\ba_t \cdot \bW_{\bM,t},
\end{equation}
yielding the updated mask map $\bM^{n+1}$.

\subsection{Refinement strategy}
Note that $\bM^0$ is defined using $\bS$ and gets wider as iteration goes on. We determine the number of refinement iterations adaptively according to the sparsity of $\bS$, \ie the number of sparse depth measurements. Suppose that $s$ depth points are valid and uniformly distributed in $\bS$. Then, the distance $\nu_s$ between adjacent sparse depth points is calculated as
\begin{equation}
    \nu_s = \sqrt{\frac{HW}{s}} - 1.
\label{eq:ideal}
\end{equation}
In addition, within a $p \times p$ window, each pixel can refine pixels within the distance of $p/2 - 1$. Therefore, to refine all pixels in an image, we set the number $N$ of iterations as
\begin{equation}
    N = \frac{\nu_s}{p/2 - 1}.
\label{eq:iterations}
\end{equation}
Since the computation in \eqref{eq:ideal} is for an ideal case, $N$ in \eqref{eq:iterations} is a lower limit for the mask to spread throughout the entire image. We, therefore, multiply $N$ by a pre-defined hyperparameter $\kappa$ for margin. We also set the minimum number of iterations to six to perform sufficient refinement even when a large number of sparse depths are given.

In MSPN, sparse depth points and refined pixels are treated similarly after iterations. This is less effective when a large number of sparse depths are provided. Hence, we use two MSPN layers and initialize the mask to $\bM^0$ as shown in Figure~\ref{fig:overview}. The second MSPN refines the depth map by reusing the accurate $\bS$ with the mask initialization. This helps to reduce the adverse impacts of monocular depth errors around the sparse depths. Empirically, we fix its iteration number to six. To summarize, two MSPN layers are configured in series and perform refinement for $N$ and six iterations, respectively.

\subsection{Loss functions}
For NYUv2 \cite{silberman2012indoor}, we use both L1 and L2 losses for training. Let $d_i$ and $\hat{d}_i$ denote the $i$th depths in the ground-truth depth map $\bD$ and a predicted depth map $\hat{\bD}$, respectively. Then, the loss function is defined as
\begin{equation}
\cL(\hat{\bD}, \bD)= \frac{1}{|\bD|}\sum_{\sigma=1}^2\sum_{i}|\hat{d}_i - d_i|^\sigma
\label{eq:L1L2}
\end{equation}
where $|\bD|$ denotes the number of valid pixels in $\bD$.

For KITTI \cite{geiger2012we}, we use the scale-invariant logarithmic loss function, which is widely used in monocular depth estimation \cite{bhat2021adabins, yuan2022newcrfs}. It is defined as
\begin{equation}
    \cL_{SI}(\hat{\bD}, \bD)=\alpha \sqrt{\frac{1}{|\bD|}\sum_{i}{{e_i}^2}-\frac{\lambda}{|\bD|^2}(\sum_{i}{e_i})^2}
\label{eq:SILog}
\end{equation}
where $e_i=\log \hat{d}_i - \log d_i$. As in \cite{bhat2021adabins, yuan2022newcrfs}, we set $\alpha=10$ and $\lambda=0.85$.

\section{Experiments}
\label{sec:experiments}
\subsection{Datasets}
\textbf{NYUv2} \cite{silberman2012indoor}: It contains 464 indoor scenes captured by Kinect v1 and provides 120K and 654 images for training and testing, respectively. As done in \cite{ma2018sparse, park2020non, zhang2023completionformer}, we use the uniformly sampled 50K images for training. Also, as in \cite{silberman2012indoor}, we fill in missing depths using the colorization scheme \cite{levin2004colorization}. The original images of size $640 \times 480$ are bilinearly downsampled by a factor of $\frac{1}{2}$ and then center-cropped to $304 \times 228$. Sparse depths are randomly sampled from the ground truth depth map.

\noindent\textbf{KITTI Depth Completion (DC)} \cite{geiger2012we}: It contains outdoor scenes captured by HDL-64E. Since the depth maps obtained by HDL-64E are sparse, the measurements are used as input, and the ground truth is generated by combining the information in 11 consecutive frames. In this work, we use the 10K subset in \cite{zhang2023completionformer} for training and do the test on a 1K validation set. As in \cite{park2020non, zhang2023completionformer}, the original images are bottom-center-cropped to $240 \times 1216$.

\subsection{Evaluation metrics}
We adopt the top three metrics in Table \ref{tb:eval_metric} for NYUv2 and the bottom four for KITTI. For RMSE, NYUv2 uses meters, and KITTI uses millimeters.

\begin{table}[!h]
    \centering
    \small
    \renewcommand{\arraystretch}{1.3}
    \setlength{\tabcolsep}{7pt}
    \caption
    {
        Evaluation metrics for estimated depth maps. Here, $|\bD|$ denotes the number of valid pixels in a depth map $\bD$, $d_i$ is the $i$th valid depth in $\bD$, and $\hat{d}_i$ is an estimate of $d_i$.
    }
    \vspace{0.1cm}
    \begin{tabular}{l l}
    \toprule
    REL & $ \frac{1}{|\bD|}\sum_{i} \vert \hat{d}_i-d_i \vert /d_i$\\
    \midrule
    $\delta_k$ & \% of $d_i$ that satisfies
    $\max\!\left\{\frac{\hat{d}_i}{d_i},\frac{d_i}{\hat{d}_i}\right\}<k$\\
    \midrule
    RMSE (m, mm) & $\frac{1}{|\bD|}\big(\sum_{i}(\hat{d}_i-d_i)^2\big)^{0.5}$\\
    \midrule
    iRMSE (1/km) & $\frac{1}{|\bD|}\big(\sum_{i}(\frac{1}{\hat{d}_i}-\frac{1}{d_i})^2\big)^{0.5}$\\
    \midrule
    MAE (mm) & $ \frac{1}{|\bD|}\sum_{i} \vert \hat{d}_i-d_i \vert$\\
    \midrule
    iMAE (1/km) & $ \frac{1}{|\bD|}\sum_{i} \vert \frac{1}{\hat{d}_i}-\frac{1}{d_i} \vert$\\
    \bottomrule
    \end{tabular}
    \label{tb:eval_metric}
    \vspace{-0.3cm}
\end{table}

\begin{figure*}[!t]
  \centering
   \includegraphics[width=\linewidth]{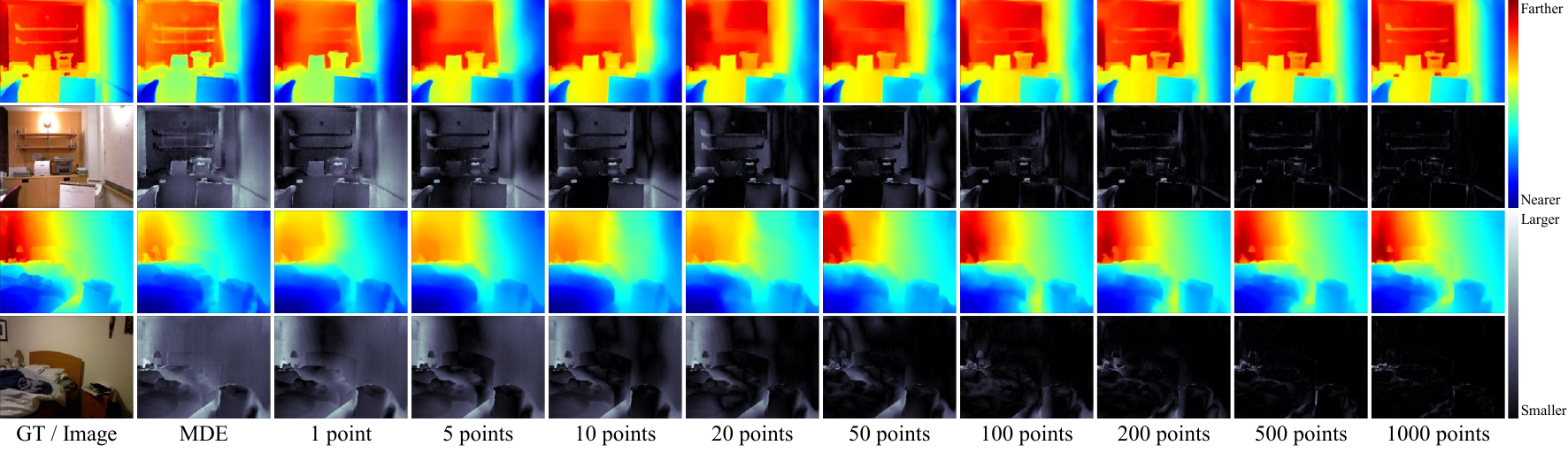}
   \caption{SDR results on NYUv2. For each depth map, the corresponding error map is provided below, in which brighter pixels represent larger errors.}
   \label{fig:qualitative_nyu}
   \vspace*{-0.2cm}
\end{figure*}

\begin{figure}[!t]
   \vspace{-0.3cm}
  \centering
   \includegraphics[width=\linewidth]{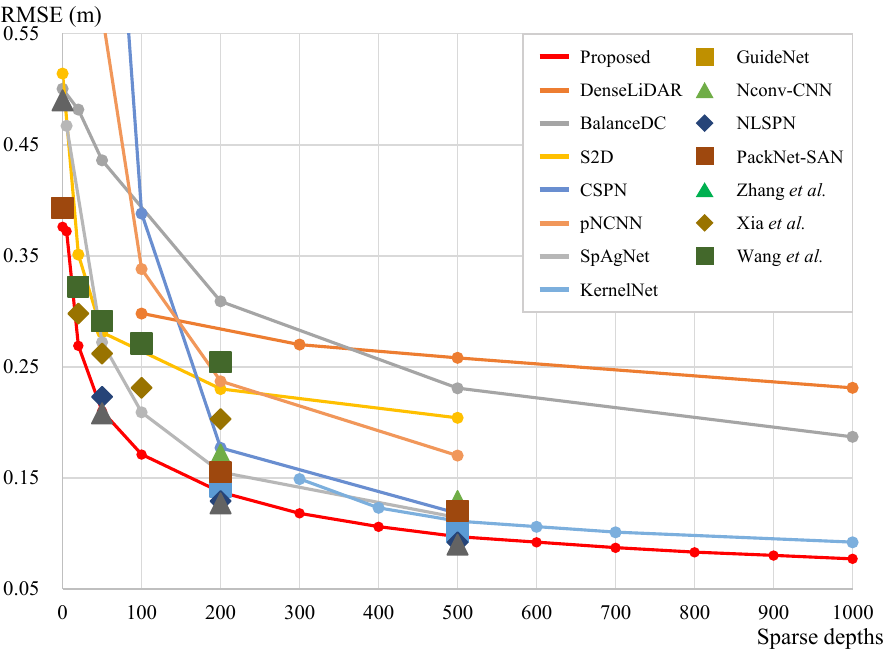}
   \caption{Comparison of the SDR performances on NYUv2.}
    \label{fig:SDR_perf_NYU}
   \vspace{-0.3cm}
\end{figure}

\subsection{Implementation details}
\noindent
\textbf{Network architecture:}
In Figure~\ref{fig:overview} (b), we adopt PVT-Base \cite{wang2021pyramid} as the encoder of the guidance network. The encoder takes the input of spatial resolution $H \times W$ and yields the output of spatial resolution $H/16 \times W/16$ with 512 channels. The decoder consists of five blocks; each upsamples its input feature using $3 \times 3$ transposed convolution, followed by layer normalization, ReLU activation, and a NAF block \cite{chen2022simple}. To the last four decoder blocks, the encoder features are skip-connected via concatenation. The guidance network outputs $\bG$ of resolution $H \times W$ with 64 channels.

\noindent
\textbf{Training:}
We train models with the AdamW optimizer \cite{loshchilov2017decoupled} with an initial learning rate of $10^{-3}$, weight decay of $10^{-2}$, $\beta_1 = 0.9$, and $\beta_2 = 0.999$. The batch size per GPU is set to 24 and 16 on NYUv2 and KITTI, respectively, using gradient accumulation. We use a $13 \times 13$ window for MSPN. For NYUv2, the model is trained for 36 epochs, and the learning rate is halved after 18, 24, and 30 epochs. For KITTI, the model is trained for 60 epochs, and the learning rate is halved after 30, 36, 42, 48, and 54 epochs. The number of iterations $N$ in (\ref{eq:iterations}) is weighted by $\kappa=2$.

For NYUv2, the number of sparse depth points is uniformly sampled between 10 and 1000. In the case of KITTI \cite{geiger2012we}, each input lidar scan has 64 lines. To simulate a real-world case and to compare with other methods, we sample lines instead of sampling each sparse depth point randomly. The number of sampled lines is set evenly between 4 and 64. To determine the number of iterations, we compute the average number of sparse depth pixels per line and use it to calculate $N$ in (\ref{eq:iterations}).

\begin{figure}[!t]
   \vspace{-0.3cm}
    \centering
   \includegraphics[width=\linewidth]{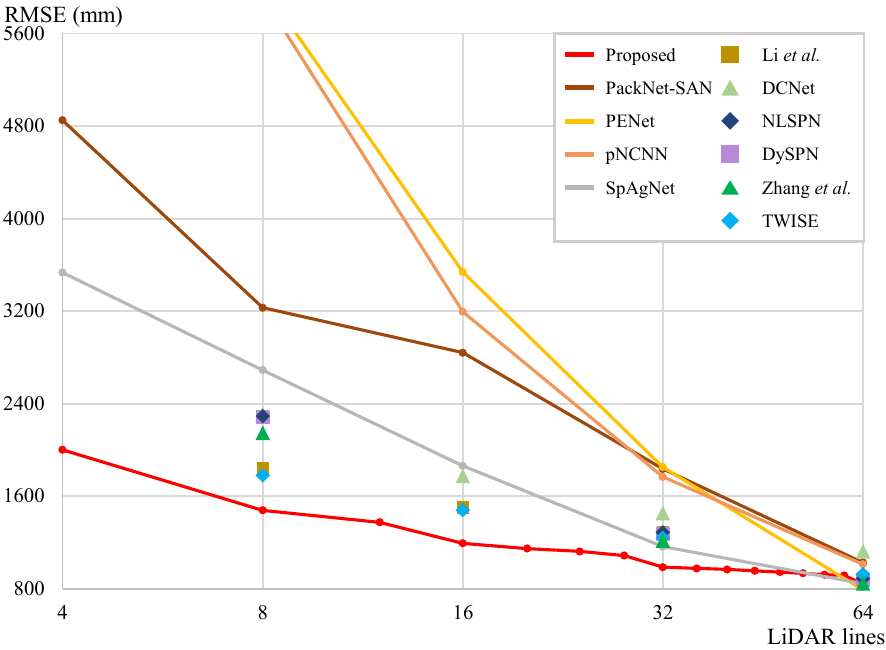}
   \caption{Comparison of the SDR performances on the KITTI validation set.}
   \label{fig:SDR_perf_KITTI}
   \vspace{-0.3cm}
\end{figure}

\begin{figure*}[!t]
  \centering
   \includegraphics[width=\linewidth]{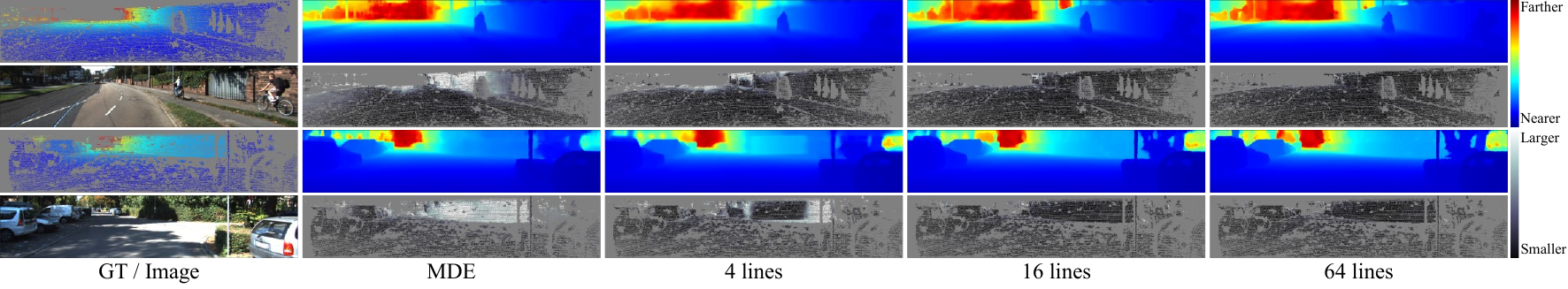}
   \caption{SDR results on KITTI. For each depth map, the corresponding error map is provided below, in which brighter pixels represent larger errors.}
   \label{fig:qualitative_kitti}
   \vspace{-0.3cm}
\end{figure*}

\vspace{-0.1cm}
\subsection{Sparsity-adaptive depth refinement}
We evaluate the SDR performances of the proposed MSPN with other depth completion algorithms \cite{wang2019plug, gu2021denselidar, yoon2020balanced, ma2018sparse, eldesokey2020uncertainty, conti2023sparsity, liu2021learning, tang2020learning, guizilini2021sparse, park2020non, eldesokey2019confidence, zhang2023completionformer, xia2020generating, hu2021penet, li2020multi, imran2019depth, imran2021depth, lin2022dynamic}. Figures~\ref{fig:SDR_perf_NYU} and \ref{fig:SDR_perf_KITTI} compare the RMSE performances according to the number of sparse depths on NYUv2 and KITTI, respectively. In Figures~\ref{fig:SDR_perf_NYU} and \ref{fig:SDR_perf_KITTI}, a solid line indicates that a single model is evaluated for various numbers of sparse depths. On the contrary, each symbol means that a separate model is trained and evaluated for a fixed number of sparse depths. The following observations can be made from Figures \ref{fig:SDR_perf_NYU} and \ref{fig:SDR_perf_KITTI}:

\begin{itemize}
\itemsep0mm
\item By comparing the solid lines in Figure \ref{fig:SDR_perf_NYU}, we see that the proposed MSPN outperforms all the other methods for all numbers of sparse depths on NYUv2.
\itemsep0mm
\item Specifically, some methods are specialized for many sparse depths, and their performances degrade significantly with fewer sparse depths. Conversely, some are specialized for few sparse depths, and their performances improve marginally with more sparse depths.
\itemsep0mm
\item On the other hand, the proposed MSPN exhibits similar performances to those symboled methods \cite{park2020non, zhang2023completionformer}, trained for specific numbers of sparse depths. This indicates that MSPN yields robust results regardless of the number of sparse depths.
\itemsep0mm
\item In Figure \ref{fig:SDR_perf_KITTI}, MSPN significantly outperforms the other methods on KITTI when less than 64 lines are available.
\itemsep0mm
\item For KITTI, methods specialized for a certain number of LiDAR lines do not perform well with a small number of lines. On the contrary, MSPN utilizes monocular depth estimation results to perform depth completion effectively, regardless of the number of lines.
\itemsep0mm
\item Overall, MSPN yields more reliable depth maps for varying numbers of sparse depths than the conventional algorithms on both indoor and outdoor images. This indicates that MSPN is more suitable for real-world applications.
\end{itemize}
Figures \ref{fig:qualitative_nyu} and \ref{fig:qualitative_kitti} show SDR results with varying numbers of sparse depths. We see that depth maps can be improved with just a single sparse depth, and errors are reduced when more sparse depths become available.

\subsection{Ordinary depth completion}
Although the primary focus of MSPN is SDR, we also evaluate the performance of MSPN under an ordinary depth completion scenario. For this ordinary depth completion, we add another decoder head to the guidance network in order to predict an initial depth map and do not use the monocular depth estimator, as in previous works \cite{park2020non, lin2022dynamic, zhang2023completionformer}.

\begin{table}[!t]
    \scriptsize
    \setlength{\tabcolsep}{14.8pt}
    \caption{Comparison of ordinary depth completion results on NYUv2. In each test, the best result is \textbf{boldfaced}.}
    \centering
    \begin{tabular}{@{\hskip 1em}l|ccc@{\hskip 1em}}
    \toprule
    Method & RMSE$(\downarrow)$ & REL$(\downarrow)$ & $\delta_{1.25} (\uparrow)$\\
    \midrule
    S2D~\cite{ma2018sparse} & 0.204 & 0.043 & 97.8\\
    GuideNet~\cite{tang2020learning} & 0.101 & 0.015 & 99.5\\
    PackNet-SAN~\cite{guizilini2021sparse} & 0.120 & 0.019 & 99.4\\
    TWISE~\cite{imran2021depth} & 0.097 & 0.013 & \textbf{99.6}\\
    NLSPN~\cite{park2020non} & 0.092 & \textbf{0.012} & \textbf{99.6}\\
    RigNet~\cite{yan2022rignet} & 0.090 & 0.013 & \textbf{99.6}\\
    DySPN~\cite{lin2022dynamic} & 0.090 & \textbf{0.012} & \textbf{99.6}\\
    Zhang~\etal~\cite{zhang2023completionformer} & 0.090 & \textbf{0.012} & -\\
    \midrule
    Proposed & \textbf{0.089} & \textbf{0.012} & \textbf{99.6}\\
    \bottomrule
    \end{tabular}
    \label{tb:performance_NYU_fixed}
\end{table}

\begin{figure}[!t]
   \vspace{-0.1cm}
    \centering
   \includegraphics[width=\linewidth]{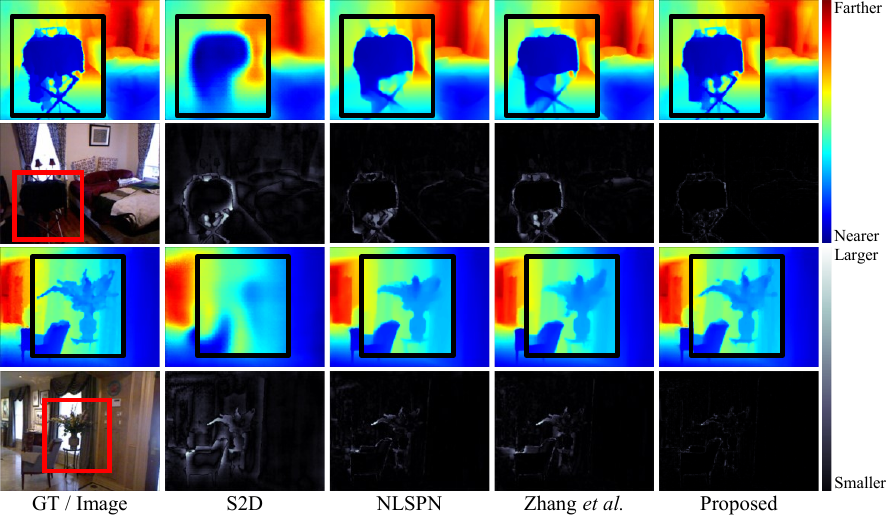}
   \caption{Qualitative comparison of ordinary depth completion results on NYUv2. }
   \label{fig:qualitative_NYU_ordinary}
   \vspace{-0.3cm}
\end{figure}

We train and test our model using a fixed amount of sparse depths. For NYUv2, we use randomly sampled 500 sparse depths from the ground truth and train the network for 72 epochs. For KITTI, we train models specialized for 16 and 64 LiDAR lines, respectively, for 72 epochs. For a fair comparison on KITTI, we use the 10k subset for training provided by \cite{zhang2023completionformer}.

Tables~\ref{tb:performance_NYU_fixed} and \ref{tb:performance_kitti_lines} compare the performances on NYUv2 and KITTI, respectively. We see that the proposed MSPN provides state-of-the-art performances on ordinary depth completion as well. Figure \ref{fig:qualitative_NYU_ordinary} qualitatively compares the results with \cite{ma2018sparse, park2020non, zhang2023completionformer} on NYUv2. We see that MSPN fills in challenging regions with fine details more effectively.

\subsection{Analysis}
\begin{figure}[!t]
  \centering
   \includegraphics[width=\linewidth]{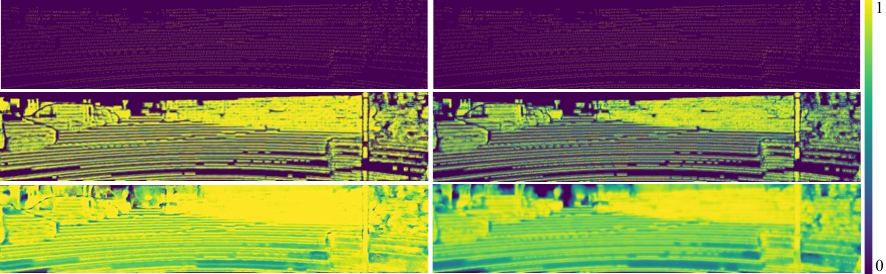}
   \caption{Illustration of the mask update process of MSPN on KITTI. The three rows represent the input, the first masks, and the last masks, respectively. The two columns indicate the masks of the first and second MSPN layers, respectively.}
   \label{fig:mask_update_KITTI}
    \vspace{-0.2cm}
\end{figure}

\begin{figure}[!t]
    \centering
   \includegraphics[width=\linewidth]{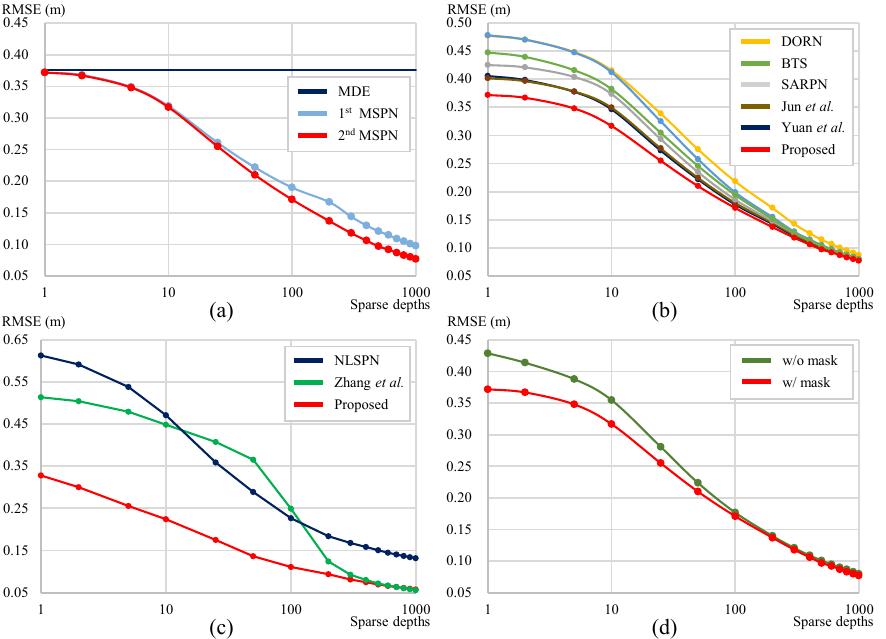}
   \caption{(a) SDR results of each MSPN layer. (b) SDR results using different monocular depth estimators. (c) Cross-dataset evaluation performance on SUN RGB-D. (d) Ablation study of the mask update process in MSPN.}
    \label{fig:analysis}
    \vspace{-0.1cm}
\end{figure}

\noindent
\textbf{MSPN:} Figure \ref{fig:mask_update_KITTI} shows the mask update process of MSPN using 16 LiDAR lines on KITTI. We see that the mask values at the second MSPN layer are less widened than those of the first MSPN layer. This indicates that the two MSPN layers play different roles --- The first layer roughly refines an entire image, and the second layer intensively refines near sparse depths. Figure \ref{fig:analysis} (a) shows the RMSE performances of each MSPN layer on NYUv2. Since the role of the second layer is to refine nearby regions of sparse depths, it has a higher gain as more sparse depths are provided.

\begin{table}[!t]
    \scriptsize
    \setlength{\tabcolsep}{4.7pt}
    \caption{Comparison of ordinary depth completion results on the KITTI validation set. In each test, the best result is \textbf{boldfaced}.}
    \centering
    \begin{tabular}{c|l|ccccc}
    \toprule
    Lines & Method & RMSE$(\downarrow)$ & MAE$(\downarrow)$ & iRMSE$(\downarrow)$ & iMAE$(\downarrow)$\\
    \midrule
    \multirow{5}{*}{16} & NLSPN~\cite{park2020non} & 1288.9 & 377.2 & 3.4 & 1.4\\
    & DySPN~\cite{lin2022dynamic} & 1274.8 & 366.4 & 3.2 & 1.3\\
    & Zhang~\etal~\cite{zhang2023completionformer} & 1218.6 & \textbf{337.4} & 3.0 & \textbf{1.2}\\
    \cmidrule{2-6}
    & Proposed & \textbf{1212.7} & 341.8 & \textbf{2.6} & \textbf{1.2}\\
    \midrule
    \multirow{5}{*}{64} & NLSPN~\cite{park2020non} & 889.4 & 238.8 & 2.6 & 1.0\\
    & DySPN~\cite{lin2022dynamic} & 878.5 & 228.6 & 2.5 & 1.0\\
    & Zhang~\etal~\cite{zhang2023completionformer} & 848.7 & \textbf{215.9} & 2.5 & \textbf{0.9}\\
    \cmidrule{2-6}
    & Proposed & \textbf{835.7} & 218.5 & \textbf{2.1} & \textbf{0.9}\\
    \bottomrule
    \end{tabular}
    \label{tb:performance_kitti_lines}
\end{table}

\noindent
\textbf{Generalization:} We assess the generalization capability of the proposed SDR. In this test, the guidance network and MSPN, trained for Jun \etal \cite{jun2023versatile}, are not fine-tuned.
In Figure \ref{fig:analysis} (b), we use other off-the-shelf networks, \cite{fu2018deep, lee2019big, chen2019structure, jun2022depth, yuan2022newcrfs} for monocular depth estimator and evaluate the SDR performances. We observe similar trends for various MDEs, which indicates that the proposed SDR framework provides robust results without being sensitive to the adopted monocular depth estimators. Figure \ref{fig:analysis} (c) compares the cross-dataset evaluation performance of SDR on RealSense split of SUN RGB-D \cite{song2015sun} with \cite{park2020non} and \cite{zhang2023completionformer}. We observe that the proposed SDR provides robust performance on unseen cameras.

\noindent
\textbf{Ablation study:} To assess the impact of the mask update process in MSPN, we remove the mask and its update process in MSPN. Figure \ref{fig:analysis} (d) compares the SDR results of this ablated setting on NYUv2. We see that the ablated setting severely degrades the SDR results, especially with a small number of sparse depths. This indicates that the mask update plays a crucial role in MSPN.

\begin{figure}[!t]
    \centering
   \includegraphics[width=\linewidth]{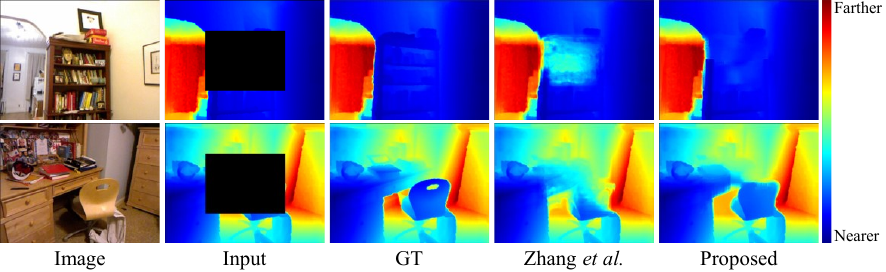}
   \caption{Qualitative comparison of depth hole filling results.}
    \label{fig:qualitative_NYU_hole_filling}
    \vspace{-0.4cm}
\end{figure}

\noindent
\textbf{Depth hole filling:} We evaluate the depth filling performances of MSPN for large areas with no sparse depths. To this end, we mask the center $114 \times 152$ region of a $228 \times 304$ ground-truth depth map and train the network to restore the masked region. We compare the results of the proposed MSPN with Zhang \etal \cite{zhang2023completionformer} in Table \ref{tb:performance_hole_filling} and Figure \ref{fig:qualitative_NYU_hole_filling}. Note that MSPN recovers missing regions more faithfully.

\begin{table}[!t]
    \scriptsize
    \setlength{\tabcolsep}{4.2pt}
    \renewcommand{\arraystretch}{1.1}
    \caption{Comparison of depth hole filling results on NYUv2. The performance is evaluated only on masked areas.}
    \centering
    \begin{tabular}{l|ccccc}
    \toprule
    Method & RMSE$(\downarrow)$ & REL$(\downarrow)$ & $\delta_{1.25} (\uparrow)$ & $\delta_{1.25^2} (\uparrow)$ & $\delta_{1.25^3} (\uparrow)$\\
    \midrule
    Zhang~\etal~\cite{zhang2023completionformer} & 0.348 & 0.073 & 0.925 & 0.982 & 0.995\\
    MSPN & \textbf{0.325} & \textbf{0.064} & \textbf{0.936} & \textbf{0.986} & \textbf{0.996}\\
    \bottomrule
    \end{tabular}
    \label{tb:performance_hole_filling}
    \vspace{-0.1cm}
\end{table}

\section{Conclusions}
In this paper, we proposed the sparsity-adaptive depth refinement (SDR) framework for real-world depth completion. First, a monocular depth estimator generates an initial depth map. Then, the guidance network generates guidance features using the initial depth map, the input image, and sparse depth measurements. Finally, the proposed MSPN gradually refines the initial depth map using the guidance features by propagating sparse depth points to the entire depth map. Extensive experiments demonstrated that the proposed MSPN provides excellent results on both SDR and conventional depth completion scenario.

\section*{Acknowledgements}
\vspace{-0.2cm}
This work was supported by the National Research Foundation of Korea (NRF) grants funded by the Korea government (MSIT) (No. NRF-2021R1A4A1031864 and NRF-2022R1A2B5B03002310).

{
    \small
    \bibliographystyle{ieeenat_fullname}
    \bibliography{1076}
}


\end{document}